\documentclass[conference]{IEEEtran}
\IEEEoverridecommandlockouts
\usepackage{cite}
\usepackage{amsmath,amssymb,amsfonts}
\usepackage{graphicx}
\usepackage{textcomp}
\usepackage{xcolor}
\usepackage{tikz}
\usepackage{bm} 
\usepackage{algorithm}
\usepackage{algpseudocode}  
\usepackage{algorithmicx}   
\def\BibTeX{{\rm B\kern-.05em{\sc i\kern-.025em b}\kern-.08em
    T\kern-.1667em\lower.7ex\hbox{E}\kern-.125emX}}
\begin{document}

\title{Worst-Case Morphs: a Theoretical and a Practical Approach
}

\author{\IEEEauthorblockN{1\textsuperscript{st} Una M. Kelly}
\IEEEauthorblockA{\textit{Data Management and Biometrics} \\
\textit{University of Twente}\\
Enschede, The Netherlands \\
u.m.kelly@utwente.nl}
\and
\IEEEauthorblockN{2\textsuperscript{nd} Luuk Spreeuwers}
\IEEEauthorblockA{\textit{Data Management and Biometrics} \\
\textit{University of Twente}\\
Enschede, The Netherlands \\
l.j.spreeuwers@utwente.nl}
\and
\IEEEauthorblockN{3\textsuperscript{rd} Raymond}
\IEEEauthorblockA{\textit{Data Management and Biometrics} \\
\textit{University of Twente}\\
Enschede, The Netherlands \\
r.n.j.veldhuis@utwente.nl}
}

\maketitle

\begin{abstract}
Face Recognition (FR) systems have been shown to be vulnerable to morphing attacks. We examine exactly how challenging morphs can become. By showing a worst-case construction in the embedding space of an FR system and using a mapping from embedding space back to image space we generate images that show that this theoretical upper bound can be approximated if the FR system is known. The resulting morphs can also succesfully fool unseen FR systems and are useful for exploring and understanding the weaknesses of FR systems. Our method contributes to gaining more insight into the vulnerability of FR systems.
\end{abstract}

\begin{IEEEkeywords}
Biometrics, Morphing Attack Detection, Face Recognition, Vulnerability of Biometric Systems
\end{IEEEkeywords}

\section{Introduction}
Several publications have shown that Face Recognition (FR) systems and humans are vulnerable to morphing attacks. A morph is an image that contains sufficient identity information from the faces of two different individuals for it to be accepted as a match when it is compared to an image of either contributing identity. This can lead to security risks in e.g. border control, since a criminal could travel using the identity document of an accomplice. Several methods to address such attacks have been proposed. In most cases, the morphs on which Morphing Attack Detection (MAD) methods are trained and tested are generated in-house, and are usually created by detecting corresponding landmarks in the faces, warping the images to an average geometry and then blending the pixel values \cite{FFM14, Ro17, MW18}.

The fact that MAD methods are often tested on datasets with similar characteristics as the sets used for training them may lead to overfitting to certain dataset-specific characteristics - especially if only one morphing algorithm was used \cite{SRB18,8553018}. In many countries someone applying for an identity document can provide their own printed passport photo, or even upload a photo digitally (e.g. Ireland, Estonia). This means that before a passport photo is stored in an electronic Machine Readable Travel Document (eMRTD), it may have been printed and scanned or digitally manipulated.

There are several other factors that affect how morphed images differ, leading to variation in the performance of MAD methods and vulnerability of FR systems. These include the morphing algorithm used, which landmark detector was used, or whether landmarks were selected manually, printing and scanning, the pairs of images selected for morphing, etc. There may be other tools for generating morphs of which the research community is not yet aware. Therefore, it is necessary to explore other potential methods and better understand the weaknesses of FR systems. Benchmarks for validating MAD methods - such as \cite{raja2020morphing_short,NIST,BOEP} - can be extended and improved by evaluating morphs generated with different methods. We contribute to this active field of research by providing a new morphing method, and by showing that there is an upper limit on how difficult morphs can become.

We investigate exactly how challenging morphs can in theory and in practice become for FR systems. Given an FR system and two images of two different people, the \textit{worst-case} morph is the image that is most similar to both images, according to this FR system. We train a neural network to generate images that approximate such worst-case morphs and examine how well our approximations of worst-case morphs can fool an FR system, and whether this extends to other systems.  Our results show it is possible to approximate the worst case when the (weights of the) FR system are known, which means that morphing attacks can become even more challenging than landmark- or GAN-based morphs. In black box scenarios, when the network weights are not accessible, our approximations of the worst case are less challenging, but still pose a significant threat.

Our proposed system differs from GAN-based morphs \cite{9404267,VZ20} since GANs are trained to generate images from an embedding space (different from the embedding space of an FR system) that look like real images, while we directly work with the embedding space of an FR system and its associated (dis)similarity measure to approximate the most challenging morph, and we train without the adversarial constraint of GANs.

Our main contributions are: a theoretical framework that when given an FR system and two images can be used to define a \textit{worst-case} embedding, a practical demonstration of finding images that (approximately) correspond to this worst-case embedding, showing that these can be significantly more challenging than landmark- or GAN-based morphs, examining the vulnerability of other FR systems, and providing a new method for generating morphs that leads to more variation in morph datasets, which may help researchers more accurately validate MAD methods in future work and uncover vulnerabilities of FR systems. 

\section{Background}
Research on variation in morphing algorithms includes: post-processesing landmark-based morphs to mask effects caused by the morphing process \cite{8897214}, a model to simulate the effects of printing and scanning \cite{FFM21} and considering the influence of ageing on morphing attacks \cite{9304856}.
GANs were used in \cite{VZ20, 9404267, arxiv.2108.09130} in an attempt to create a different type of morph, which was shown to be able to fool FR systems, if not as consistently as landmark-based morphs. Since GANs are difficult to train more stable methods would be useful \cite{8253599}. The lack of variation in morphing techniques is adressed in \cite{SKR20}. A method for MAD is presented and evaluated on morphs created using different algorithms, which are all landmark-based. Printed-and-scanned morphs are included, but GAN morphs or other methods are not taken into consideration.

\newcommand{\Cross}{$\mathbin{\tikz [x=1.4ex,y=1.4ex,line width=.2ex, red] \draw (0,0) -- (1,1) (0,1) -- (1,0);}$}%
\newcommand{\Checkmark}{$\checkmark$}

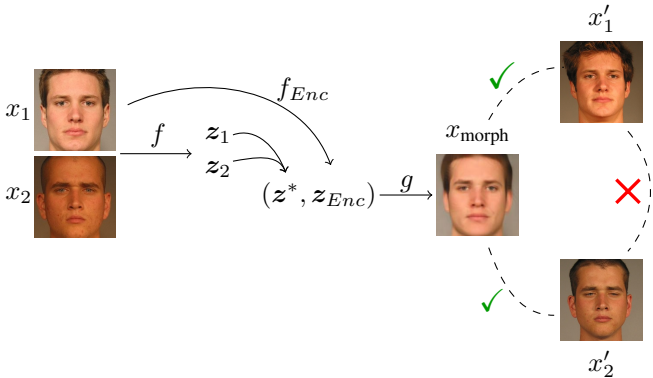
\begin{figure}[h]
	\centering
	\begin{tikzpicture}
	\filldraw[black] (-1.05,1.1) node[anchor=west,rotate=0] {$x_1$};
	\node[inner sep=0pt] (whitehead) at (0,1.13)
	{\includegraphics[width=0.05\paperwidth]{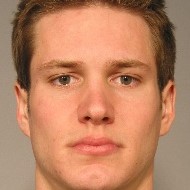}};
	
	\filldraw[black] (-1.05,0.0) node[anchor=west,rotate=0] {$x_2$};
	\node[inner sep=0pt] (whitehead) at (0,-0.03)
	{\includegraphics[width=0.05\paperwidth]{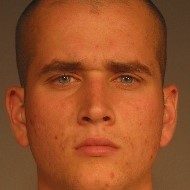}};
	
	\draw[->,black] (0.6,0.55) to [out=0,in=180] (1.55,0.55);
	\filldraw[black] (0.875,0.8) node[anchor=west,rotate=0] {$f$};
	\filldraw[black] (1.6,0.8) node[anchor=west,rotate=0] {$\bm{z}_1$};
	\filldraw[black] (1.6,0.35) node[anchor=west,rotate=0] {$\bm{z}_2$};
	
	\draw[black, ->] (2.1,0.8) to [out=20,in=110] (2.8,0.3);
	\draw[black, ->] (2.1,0.45) to [out=10,in=110] (2.8,0.3);
	
	\draw[black, ->] (0.7,1.2) to [out=25,in=110] (3.4,0.4);
	\filldraw[black] (2.55,1.4) node[anchor=west,rotate=0] {$f_{Enc}$};
	
	\filldraw[black] (2.35,0.0) node[anchor=west,rotate=0] {$(\bm{z}^*,\bm{z}_{Enc})$};
	\draw[black, ->] (4.05,0.0) to (4.75,0.0);
	\filldraw[black] (4.2,0.15) node[anchor=west,rotate=0] {$g$};
	
	\node[inner sep=0pt] (whitehead) at (5.35,0)
	{\includegraphics[width=0.05\paperwidth]{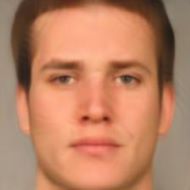}};
	\filldraw[black] (4.8,0.8) node[anchor=west,rotate=0] {$x_{\text{morph}}$};
	
	\filldraw[black] (6.7,2.35) node[anchor=west,rotate=0] {$x'_1$};	
	\node[inner sep=0pt] (whitehead) at (7.0,1.5)
	{\includegraphics[width=0.05\paperwidth]{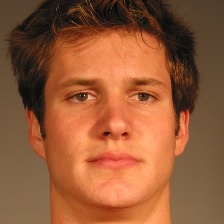}};
	
	\filldraw[black] (6.7,-2.25) node[anchor=west,rotate=0] {$x'_2$};	
	\node[inner sep=0pt] (whitehead) at (7.0,-1.4)
	{\includegraphics[width=0.05\paperwidth]{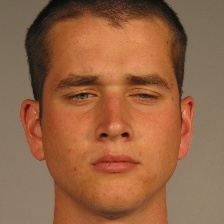}};
	
	\draw[dashed, -] (5.5,1.0) to [out=60,in=180] (6.5,1.7);
	\draw[dashed, -] (5.5,-0.7) to [out=-70,in=180] (6.37,-1.6);
	
	\filldraw[black!40!green] (5.35,1.6) node[anchor=west,rotate=0] {\Large \Checkmark};
	\filldraw[black!40!green] (5.25,-1.4) node[anchor=west,rotate=0] {\large \Checkmark};
	
	\draw[dashed, -] (7.35,0.85) to [out=-50,in=50] (7.35,-0.749);
	\filldraw[black!40!green] (7.05,0.05) node[anchor=west,rotate=0] {\Large \Cross};
	\end{tikzpicture}
	\caption{Overview of our morphing approach. Here, $x_1, x'_1$ are images of one person and $x_2, x'_2$ are images of a different person. While $x_{\text{morph}}$ visually looks like $x_1$, according to the FR system it is a succesful match with both $x'_1$ and $x'_2$. Dashed lines represent comparisons by the FR system.\label{fig:overview}}
\end{figure}

\section{Proposed System \label{approach}}
\begin{figure}[h]
	\centering
	\begin{tikzpicture}
	\draw[dashed, black, -] (0,0) -- (3,-0.5);
	
	\node at (0,0) [circle,fill,inner sep=1.5pt]{};
	\node[] at (0.2,0.2)
	{\text{$z_1$}};
	
	\node at (3,-0.5) [circle,fill,inner sep=1.5pt]{};
	\node[inner sep=0pt] (whitehead) at (3.2,-0.3)
	{\text{$z_2$}};
	
	\node at (1.5,-0.25) [circle,fill,inner sep=1.5pt]{};
	\node[inner sep=0pt] (whitehead) at (2.1,-0.0)
	{\text{$z^* = \frac{z_1+z_2}{2}$}};
	
	\node at (1.0,-0.45) [circle,fill,inner sep=1.5pt]{};
	\node[inner sep=0pt] (whitehead) at (0.9,-0.8)
	{\text{$z_{\text{lm}}$}};
	
	\node at (1.9,-0.7) [circle,fill,inner sep=1.5pt]{};
	\node[inner sep=0pt] (whitehead) at (2.1,-1.0)
	{\text{$z_{\text{GAN}}$}};
	
	\draw[black,dashed] (5.4,-0.2) circle (1cm);
	
	\node at (6.1,0.5) [circle,fill,inner sep=1.5pt]{};
	\node[] at (6.4,0.7)
	{\text{$z_1$}};
	
	\node at (5.8,-1.1) [circle,fill,inner sep=1.5pt]{};
	\node[] at (5.4,-1.5)
	{\text{$z_2$}};
	
	\node at (6.375,-0.4) [circle,fill,inner sep=1.5pt]{};
	\node[] at (7.6,-0.45)
	{\text{$z^* = \frac{z_1+z_2}{||z_1+z_2||}$}};
	
	\draw[black, -] (5.4,-0.2) -- (6.1,0.5);
	\draw[black, -] (5.4,-0.2) -- (5.8,-1.1);
	\draw[black, -] (5.4,-0.2) -- (6.375,-0.4);
	
	\node[] at (5.9,-0.6)
	{\text{$\theta$}};
	\draw[gray] (5.4,-0.2) -- (5.6,-0.65) arc [start angle=-60, delta angle=102, radius=0.5cm] -- (5.4,-0.2);
	
	\node at (6.37,-0.0) [circle,fill,inner sep=1.5pt]{};
	\node[inner sep=0pt] (whitehead) at (6.75,0.1)
	{\text{$z_{\text{lm}}$}};
	
	\node at (6.2,-0.8) [circle,fill,inner sep=1.5pt]{};
	\node[inner sep=0pt] (whitehead) at (6.6,-1.0)
	{\text{$z_{\text{GAN}}$}};
	
	\end{tikzpicture}
	\caption{\label{worstcase}The worst-case embedding $z^*$ when $d$ denotes euclidean distance (left) and angle (right). An image that maps to $z^*$ is even more challenging than a landmark- ($z_{\text{lm}}$) or GAN-based morph ($z_{\text{GAN}}$).}
\end{figure}
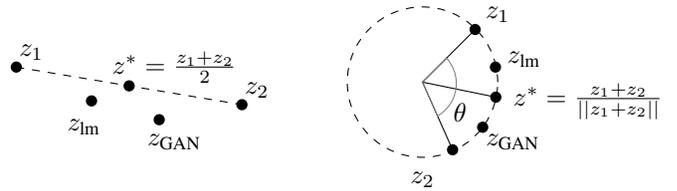

Traditionally, landmark-based morphing attempts to create images in image space that are similar to two given images. Instead, we use the embedding space of an FR system, since this should contain structured information on the similarity of the identity information present in images. Let $f$ be the function that describes an FR system's mapping from the image space $X$ to the embedding space $Z$, i.e. $f: X \rightarrow Z$ (Fig.~\ref{fig:overview}). Let $d$ be the dissimilarity score that calculated the dissimilarity of embedding pairs in $Z$. We define the \textit{worst-case embedding} for two images $x_1$ and $x_2$ as
\begin{equation}
z^* := \text{argmin}_{z \in Z} \left( \max \left[ d(z,f(x_1)), d(z,f(x_2)) \right] \right). \label{z_wc} 
\end{equation}
For example, if $S$ returns cosine similarity, then $S(z_1, z_2) =\cos(\theta)$, where $\theta$ is the angle between $z_1$ and $z_2$, see Fig.~\ref{worstcase}. In that case $z^*$ is any $z$ for which $S(z_1,z)=S(z,z_2)=\cos(\theta/2)$. If an FR system uses similarity scores, defined by a function $S$, then $S$ replaces $d$, argmin is replaced by argmax, and max is replaced by min. 

A worst-case morph is an image $x^*$ for which $f(x^*)=z^*$. We approximate $x^*$ using a decoder $D$ that maps from $Z$ to $X$, reversing the mapping of the FR system. Since $f$ is a many-to-one mapping there may be many different approximative inverse mappings for one given embedding. We choose to train $D$ to reconstruct $x_1$ when given $z^*$ as input, i.e. $D$ learns to approximate the mapping $g^*: Z \rightarrow X$, where $g^*(z^*)= x_1$ and $f(g^*(z^*))= z^*$.
$D$ generates images that to humans look like $x_1$, i.e. like \textit{one} of the input identities (e.g. the accomplice), but to the FR system are a morph of $x_1$ and $x_2$. According to the FR system the resulting image contains identity features of both accomplice and criminal.

Instead of training a decoder to generate morphed images, we could also use optimisation on an FR system. In that case, the FR latent embeddings corresponding to an image are manipulated by calculating gradients and computing adversarial perturbations. The disadvantage of learning (i.e. training a decoder or other network) over optimisation (i.e. gradient descent) is that the decoder is trained to generate many different images, while optimisation applies an adversarial perturbation tailored specifically to one image. The advantage of learning over optimisation is that there is no need for retraining or calculating image-specific gradients every time a new image is manipulated (e.g. from a new dataset).

\begin{figure}[t]
	\centering
	\begin{tikzpicture}
	\newcommand\x{1.6}
	\newcommand\w{0.07}
	\newcommand{\shiftleft}{1.0}
	\newcommand{\shiftdown}{1.6}
	\filldraw[black] (0 -0.25, 0.95) node[anchor=west] {\footnotesize{$x_1$}};
	\node[inner sep=0pt] (whitehead) at (0,0)
	{\includegraphics[width=\w\paperwidth]{img/04361d171}};
	\filldraw[black] (\x *1 +0.325 -\shiftleft, 0.95) node[anchor=west] {\footnotesize{Landmark}};
	\node[inner sep=0pt] (whitehead) at (\x,0)
	{\includegraphics[width=\w\paperwidth]{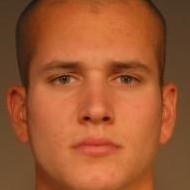}};
	\node[inner sep=0pt] (whitehead) at (\x,0 - \shiftdown)
	{\includegraphics[width=\w\paperwidth]{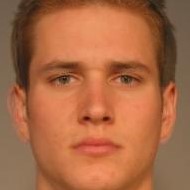}};
	\node[inner sep=0pt] (whitehead) at (\x,0- \shiftdown *2)
	{\includegraphics[width=\w\paperwidth]{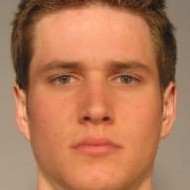}};
	\filldraw[black] (\x *2 +0.55 -\shiftleft, 0.95) node[anchor=west] {\footnotesize{GAN}};
	\node[inner sep=0pt] (whitehead) at (\x *2,0)
	{\includegraphics[width=\w\paperwidth]{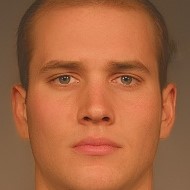}};
	\filldraw[black] (\x *3 +0.3-\shiftleft, 0.95) node[anchor=west] {\footnotesize{worst case}};
	\node[inner sep=0pt] (whitehead) at (\x *3,0)
	{\includegraphics[width=\w\paperwidth]{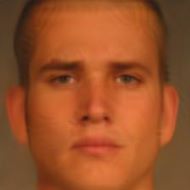}};
	\node[inner sep=0pt] (whitehead) at (\x *3,0 - \shiftdown)
	{\includegraphics[width=\w\paperwidth]{img/04361d171_04484d36_2}};
	\filldraw[black] (\x *4 +0.75 -\shiftleft, 0.95) node[anchor=west] {\footnotesize{$x_2$}};
	\node[inner sep=0pt] (whitehead) at (\x *4,0)
	{\includegraphics[width=\w\paperwidth]{img/04484d36}};
	
	\draw[line width=0.3mm, red, -] (0.787,1.12) -- (5.6,1.12) -- (5.6,-4.0) -- (0.787,-4.0) -- (0.787,1.13);
	\draw[line width=0.3mm, red, -] (0.787,1.12 - \shiftdown *3.7) -- (5.6,1.12 - \shiftdown *3.7) -- (5.6,-4.0 - \shiftdown *3.7) -- (0.787,-4.0 - \shiftdown *3.7) -- (0.787,1.13 - \shiftdown *3.7);
	
	\filldraw[black] (0 -0.25, 0.95 - \shiftdown *3.7) node[anchor=west] {\footnotesize{$x_1$}};
	\node[inner sep=0pt] (whitehead) at (0,0 - \shiftdown *3.7)
	{\includegraphics[width=\w\paperwidth]{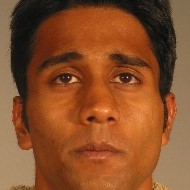}};
	\filldraw[black] (\x *1 +0.325 -\shiftleft +0.05, 0.95 - \shiftdown *3.7) node[anchor=west] {\footnotesize{Landmark}};
	\node[inner sep=0pt] (whitehead) at (\x ,0 - \shiftdown *3.7)
	{\includegraphics[width=\w\paperwidth]{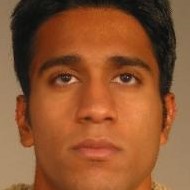}};
	\node[inner sep=0pt] (whitehead) at (\x ,0 - \shiftdown - \shiftdown *3.7)
	{\includegraphics[width=\w\paperwidth]{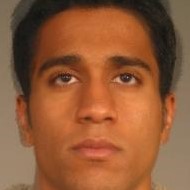}};
	\node[inner sep=0pt] (whitehead) at (\x ,0- \shiftdown *2 - \shiftdown *3.7)
	{\includegraphics[width=\w\paperwidth]{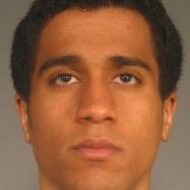}};
	\filldraw[black] (\x *2 +0.55 -\shiftleft, 0.95 - \shiftdown *3.7) node[anchor=west] {\footnotesize{GAN}};
	\node[inner sep=0pt] (whitehead) at (\x *2, 0 - \shiftdown *3.7)
	{\includegraphics[width=\w\paperwidth]{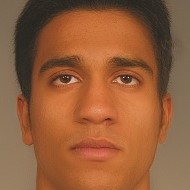}};
	\filldraw[black] (\x *3 +0.3 -\shiftleft, 0.95 - \shiftdown *3.7) node[anchor=west] {\footnotesize{worst case}};
	\node[inner sep=0pt] (whitehead) at (\x *3, 0 - \shiftdown *3.7)
	{\includegraphics[width=\w\paperwidth]{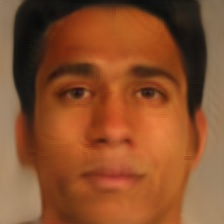}};
	\node[inner sep=0pt] (whitehead) at (\x *3, 0 - \shiftdown - \shiftdown *3.7)
	{\includegraphics[width=\w\paperwidth]{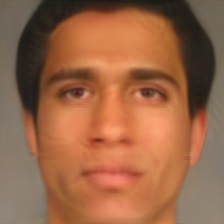}};
	\filldraw[black] (\x *4 +0.75 -\shiftleft , 0.95 - \shiftdown *3.7) node[anchor=west] {\footnotesize{$x_2$}};
	\node[inner sep=0pt] (whitehead) at (\x *4, 0 - \shiftdown *3.7)
	{\includegraphics[width=\w\paperwidth]{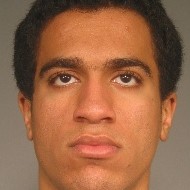}};
	
	\end{tikzpicture}
	\caption{\label{worst-case}Landmark-based morphs (spliced into $x_1$, full morph, spliced into $x_2$), GAN-based morph and worst-case approximations (with $z_{\text{Enc}}$ from $x_1$ and with $z_{\text{Enc}}$ from $x_2$).}
\end{figure}

\subsection{Morph generation}
For our experiments we use MobileFaceNet, which on LFW achieves an accuracy of 99.55\% \cite{MobileFaceNet}. For images $x_1$ and $x_2$, this FR system returns 128-dimensional embeddings $z_1=f(x_1)$ and $z_2=f(x_2)$, where $f$ describes the mapping defined by the neural network from the image space to the embedding space:
\begin{align*}
X = [0,1]^{n_c\times w\times h}  \quad \text{ and }	\quad Z = \left\{ z \in \mathbb{R}^{128}  \ \middle| \ ||z||=1 \right\},
\end{align*}
where $n_c=3$ and $w\times h=112\times 112$. It can be used as a face verification system by calculating the angle $\theta$ between $z_1$ and $z_2$. $x_1$ and $x_2$ are not accepted as a match if $\theta>t$, where $t$ is the decision threshold.
In this case, the worst-case embedding $z^*$ that minimises Eq. \ref{z_wc} is the embedding that lies on the unit hypersphere exactly between $z_1$ and $z_2$, i.e. a worst-case morph is any image $x^*$ for which $f(M^*)=z^*=\frac{z_1+z_2}{||z_1+z_2||}$. Scaling is necessary to ensure that $z^*$ lies on the unit hypersphere.

An embedding $z=f(x)$ should contain information essential to the identity of $x$, but some information such as background or expression is not relevant to identity and should therefore not be contained in $z$. This means that the FR latent embedding alone is not sufficient to reconstruct $x$, so we train a supporting encoder network that also compresses image information, including other attributes. We train the decoder network using both worst-case embeddings of the FR system and embeddings of the encoder as input.

Let $f_{\text{Enc}}$ be the mapping of the supporting encoder, i.e. $f_{\text{Enc}}(x)=z_{\text{Enc}}$. Furthermore, let $g$ denote the decoder mapping back to image space. Let $d$ be the dissimilarity score function used by the FR system to compare latent embeddings, which here is the angle between embeddings. We train $D$ to approximate $g^*$ by minimising the following loss
\begin{align} \label{eq:losses}
\mathcal{L} & = \gamma_1 \mathcal{L}_{\text{pixel}} + \gamma_2 \mathcal{L}_{\text{latent}} \nonumber \\
& = \gamma_1 \mathbb{E}_{x}\left[ ||x_{\text{morph}}-x_1||^2_2 \right] + \gamma_2 \mathbb{E}_{x}\left[d(z_{\text{morph}}, z^*)\right],
\end{align}
where $\mathcal{L}_{\text{pixel}}$ ensures that the output images are visually similar to the input images, and $\mathcal{L}_{\text{latent}}$ encourages the corresponding FR embeddings to be close to the worst case. The architecture and hyperparameters of this network can be found in the appendix. We train the network for 200 epochs, after which the losses no longer significantly decrease. 

\begin{algorithm}[t]
	\small
	\begin{algorithmic}
		\State $\theta_{D}, \theta_{\text{Enc}} \gets \text{initialize network parameters}$
		\Repeat
		\State $\bm{x}^{(1)}, \ldots, \bm{x}^{(N)}$
		\Comment{{\footnotesize Draw $N$ samples}}
		
		\State $\bm{z}^{(i)}= f(\bm{x}^{(i)}),		\ i = 1, .., N $
		\Comment{{\footnotesize Get FR embeddings}}
		
		\State $\bm{z}^{*(i)}= \frac{\bm{z}^{(i)}+\bm{z}^{(j)}}{||\bm{z}^{(i)}+\bm{z}^{(j)}||},	\ j = 2, .., N,1$
		\Comment{{\footnotesize Translate batch to get\\
		\hfill pairs for worst-case emb.}}
		
		\State $\bm{z}^{(i)}_{\text{Enc}}= f_{\text{Enc}}(\bm{x}^{(i)}),		\quad i = 1, .., N$
		\Comment{{\footnotesize Get Encoder emb.}}
		\State{$\bm{x}_{\text{morph}}^{(i)} = D(\bm{z}^{(i)}, \bm{z}_{\text{Enc}}^{(i)}),
			\ i = 1,.., N$}
		\Comment{{\footnotesize Generate morphs}}
		
		\State{$\bm{z}_{\text{morph}}^{(i)}= f(\bm{x}_{\text{morph}}^{(i)}),		\quad i = 1, .., N$}
		\Comment{{\footnotesize Get FR emb. of morphs.}}
		
		\State $\mathcal{L} = \gamma_1 \frac{1}{N} \sum_{i=0}^N \text{MSE}(\bm{x}^{(i)}, \bm{x}_{\text{morph}}^{(i)})$
		\State $ \qquad \qquad + \ \gamma_2 \frac{1}{N} \sum_{i=0}^N d(\bm{z}^{*(i)}, \bm{z}^{(i)}_{\text{morph}}) $
		\Comment{{\footnotesize Compute loss}}
		
		\State $\theta_{D} \gets \theta_{D} - \nabla_{\theta_{D}} \mathcal{L}$
		\Comment{{\footnotesize Gradient update on decoder}}
		
		\State $\theta_{\text{Enc}} \gets \theta_{\text{Enc}} - \nabla_{\theta_{\text{Enc}}} \mathcal{L}$
		\Comment{{\footnotesize Gradient update on encoder}}
		
		\Until{convergence}
	\end{algorithmic}
	\caption{\label{alg:pseudo} Training procedure. $\gamma_1=1, \gamma_2=0.1$ }
\end{algorithm}
We use a dataset of in total 21,772 facial images from \cite{FRGC} and separate them into 18,143 training and 3,629 validation images, with no overlap in identities. We create three sets of morphs using the validation set: approximations of worst-case morphs, landmark-based morphs and GAN-based morphs using MIPGAN-I \cite{9404267}.
We select pairs of similar identities and faces with neutral expression, resulting in 506 pairs for morphing. For each pair $(x_1,x_2)$ we create two worst-case approximations, three landmark morphs (one full morph and two spliced morphs), and one GAN morph, see Fig. \ref{worst-case}.

We improve each morph generated with the trained Decoder by freezing the network's weights and performing gradient descent on $\mathcal{L}_{\text{latent}} $ (Eq. 4) for $N$ iterations to improve input selection, where $z_1^* = z^* = \frac{z_1+z_2}{||z_1+z_2||}$ and $z_i^* = z_{i-1}^* - \nabla \mathcal{L}_{\text{latent}}(z_{\text{morph}})$. The improved morph is obtained by forwarding $(z_{N}^*,z_{\text{Enc}})$ through the decoder. We set $N=200$, after which the white box FR scores still decrease, but not the scores for black box FR systems.

\begin{figure*}[h]
	\includegraphics[height=0.15\paperheight]{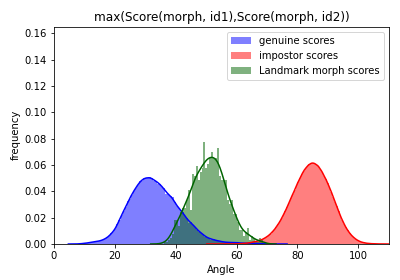}
	\includegraphics[height=0.15\paperheight]{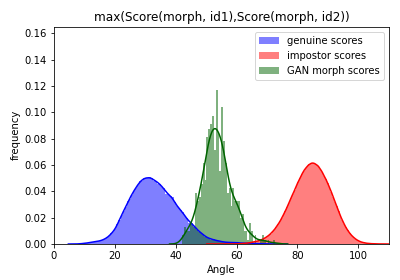}
	\includegraphics[height=0.15\paperheight]{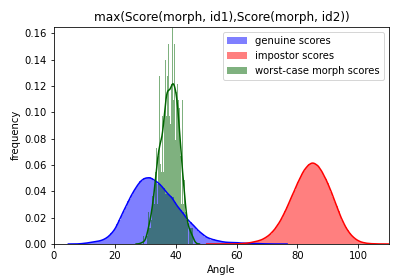}
	
	\includegraphics[height=0.15\paperheight]{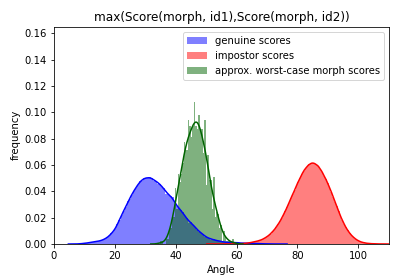}
	\includegraphics[height=0.15\paperheight]{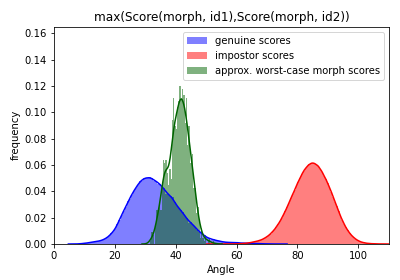}
	\caption{\label{histograms}Vulnerability of MobileFaceNet to: landmark morphs, MIPGAN-I morphs, worst-case embeddings (theoretical upper bound), approximations of worst-case morphs, and improved approximations after optimisation (200 iterations per image). Blue histograms describe genuine comparison scores, impostor comparison scores are red. For each morph we compute the angles for probe images of each contributing identity. The green histograms describe the largest of these two angles. Our morphs are extremely difficult to distinguish from genuines using FR comparison scores.}
\end{figure*}

\begin{table}
	\centering
	\caption{MMPMR (\%).}\label{tab1}
	\begin{tabular}{|l@{\hspace{0.11cm}}|l@{\hspace{0.11cm}}|l@{\hspace{0.11cm}}|l@{\hspace{0.11cm}}|l@{\hspace{0.11cm}}|l@{\hspace{0.11cm}}|}
		\hline & & & & &\vspace{-0.3cm}\\
		& Landmark	& GAN	& Worst-	& Worst-	& Improved \\
		& 			& 		& case		& case		&  worst-case \\
		& 			& 		& 			& approx.	&  approx.\\
		\hline & & & & &\vspace{-0.3cm}\\
		\textit{MobileFaceNet}		& 97.3	& 96.3	& 100.0 & 100.0 & 100.0\\
		\hline & & & & &\vspace{-0.3cm}\\
		\textit{VGG16}	(black box) & 63.0	& 65.0	& 100.0 & 33.6 & 43.1\\
		\hline & & & & &\vspace{-0.3cm}\\
		\textit{ArcFace} (black box)& 85.4	& 66.3	& 100.0 & 47.9 & 61.1\\
		\hline
	\end{tabular}
\end{table}

\section{Results \& Discussion}\label{results}
We compare the vulnerability of FR systems to landmark-based, GAN-based morphing attacks and worst-case approximations. We examine the morphs with the same deep-learning-based FR system used for training the inverse network and with two other FR systems, where we calculate the Mated Morph Presentation Match Rate (MMPMR($t$)), which is the proportion of (morphing) attacks for which both contributing identities are considered a match by the FR system when using a threshold~$t$ \cite{SNR17}. For each FR system we set $t$ such that the false non-match rate is minimal while the false match rate$<$0.1\%. 

The histograms of MobileFaceNet comparison scores are shown in Fig. \ref{histograms}. The MMPMR for the worst-case approximations is 100\%, even higher than the 97.3\% for landmark-based morphs (Tab. \ref{tab1}). This means that every worst-case approximation contains enough identity information of both contributing identities to fool the FR system. 

In Fig. \ref{histograms} we can see that the worst-case approximations are very close to the actual worst case, especially after optimising each morph separately. Table 1 shows that this does not necessarily hold for other FR systems. 
This is not entirely surprising, since our method is trained to generate approximations of worst-case morphs specifically for one FR system, while the worst-case morphs for a different FR system might be different. Nonetheless, a significant proportion of the morphs is still accepted as a match with both contributing identities. Most importantly, we have shown that it is possible to generate images that are close to the worst case, and therefore believe that researchers should not focus exclusively on landmark- and GAN-based morphs, since more challenging morphs may exist.

\section{Conclusion \& Future Work}

We introduced the concept of worst-case morphs, which are morphs that are most similar to the contributing identities according to an FR system. We trained a decoder network to generate images that approximate such worst-case morphs. These were extremely successful at fooling the FR system that was used for training and significantly more challenging than landmark- or GAN-based morphs. We also showed that generating morphs that are close to the worst case for FR systems under black box settings is still an open problem. This might be improved by minimising the distance to the worst-case embedding for several FR systems simultaneously. Since the quality of morphs depends on the image quality of genuine images, pair selection, and other factors, a good measure to compare the quality of different morphs could be measuring the distance to the worst-case embedding.

The morphs we generate may be helpful in understanding the vulnerabilities of FR systems and may also offer some insight into the robustness of MAD techniques. Since the morphs generated using the decoder network are closer to the worst case than landmark- or GAN-based morphs, even before applying optimisation to improve each morph individually, challenging morphs can be generated in a computationally inexpensive manner. Furthermore, our approach has the advantage of being more stable to train than e.g. GANs. We also studied the potential of our method for use in black box settings.

\bibliographystyle{ieee}
\bibliography{draft}

\section*{Hyperparameters}
\newcommand{\Crossblack}{$\mathbin{\tikz [x=1.4ex,y=1.4ex,line width=.2ex, black] \draw (0,0) -- (1,1) (0,1) -- (1,0);}$}%

\begin{table}[h]
	\centering
	\footnotesize
	\begin{tabular}{@{}rlll@{}}
		Operation   & Kernel       & Stride			& Size
		\vspace{-0.25cm}
		\\
		\hrulefill & & & \hrulefill
		\vspace{-0.06cm}
		\\
		$f_{\text{Enc}}(\bm{x})$ & & 	& $224 \times 224 \times 3$ (Input)\\
		\vspace{-0.55cm} 
		\\
		\hrulefill & & & \hrulefill
		\vspace{-0.05cm}                                                             \\
		5$\times$ Convolution & $5 \times 5$ & $2 \times 2$	& $7 \times 7 \times 64$
		\vspace{-0.25cm}
		\\
		\hrulefill & & & \hrulefill
		\vspace{-0.06cm}
		\\
		$g(\bm{z})$ & & 	& $1 \times 1 \times 128$ (Input)\\
		\vspace{-0.55cm} 
		\\
		\hrulefill & & & \hrulefill\\
		Fully connected	  & - 			 & -      		& $1 \times 1 \times 512$ \\ 
		Fully connected	  & - 			 & -      		& $7 \times 7 \times 448$ \\ 
		Concat. with $f_{\text{Enc}}(\bm{x})$& &     	&$7 \times 7 \times 512$
		\\
		4$\times$ Upsample \& Conv. & $3 \times 3$ & $1 \times 1$ & $112 \times 112 \times 32$ \\
		Conv. & $3 \times 3$ & $1 \times 1$ & $112 \times 112 \times 32$ \\
		Upsample \& Conv. & $3 \times 3$ & $1 \times 1$ & $224 \times 224 \times 3$
		\vspace{-0.25cm}
		\\
		\hrulefill
		\vspace{-0.05cm}\\
		Optimizer		& \multicolumn{3}{@{}l@{}}{Adam($\alpha = 10^{-4}$, $\beta_1 = 0.0$, $\beta_2 = 0.9)$}  \\
		Batch size		& \multicolumn{3}{@{}l@{}}{64} \\	
		Bias			& \multicolumn{3}{@{}p{5.5cm}@{}}{False, except in the last layer of the decoder, where it is untied.} \\	
		BatchNorm		& \multicolumn{3}{@{}p{5.5cm}@{}}{After each convolution.} \\		
		Weight init. 	& \multicolumn{3}{@{}l@{}}{Isotropic gaussian ($\mu = 0$, $\sigma = 0.01$)} \\	
		Untied bias init.& \multicolumn{3}{@{}p{5.5cm}@{}}{$\text{Sigmoid}^{-1}(\frac{1}{n} \sum_{i=1}^n \bm{x}_i$), the average of $\{\bm{x}_1,\dots, \bm{x}_n\}$.} \\
		
		Nonlinearity	& \multicolumn{3}{@{}p{5.5cm}@{}}{Leaky ReLU of slope 0.02, except for the last convolution in $g$, which is followed by a sigmoid activation function.}
	\end{tabular}
\end{table}

\end{document}